\documentclass{ifacconf}

\usepackage{graphicx}      
\usepackage{natbib}        
\usepackage{amsmath}
\usepackage{amssymb}
\usepackage{tikz}
\usepackage{diagbox}
\usepackage{array}
\usepackage{comment}
\usepackage{booktabs}

\begin{document}
\begin{frontmatter}

\title{Multi Agent Switching Mode Controller for Sound Source Localization} 

\thanks[footnoteinfo]{This work is partly supported by the Italian PNRR project {\it SMS2MARIE: Sistemi multi-agente e Multi Sensore per il Monitoraggio, la MAppatura e la Ricerca in aree In condizioni critiche e durante Emergenze}; Grant Agreement n.PTSL-SD1701255103 CUP: C99J24000240008 Programma “Future Artificial Intelligence Research” - FAIR (Codice PE00000013), PNRR - MISSIONE 4 “Istruzione e ricerca”, COMPONENTE 2 “Dalla ricerca all’impresa”, Investimento 1.3 - bando a cascata Spoke 7. 
\
Corresponding author: Marcello Sorge, \tt{marcello.sorge@unipd.it}.}

\author[First]{Marcello Sorge} 
\author[First]{Nicola Cigarini}
\author[Second]{Riccardo Lorigiola}
\author[Third,Second]{Giulia Michieletto}
\author[First]{Andrea Masiero}
\author[Second,Fourth]{Angelo Cenedese}
\author[First]{Alberto Guarnieri}

\address[First]{Department of Land, Environment, Agriculture and Forestry, 
   University of Padova, Legnaro (PD), Italy.}
\address[Second]{Department of Information Engineering, University of Padova, Padova, Italy.}
\address[Third]{Department of Management and Engineering, 
   University of Padova, Vicenza, Italy.}
\address[Fourth]{Department of Industrial Engineering, University of Padova, Padova, Italy.}

\begin{abstract}                
Source seeking is an important topic in robotic research, especially considering sound-based sensors since they allow the agents to locate a target even in critical conditions where it is not possible to establish a direct line of sight. In this work, we design a multi-agent switching mode control strategy for acoustic-based target localization. Two scenarios are considered: single source localization, in which the agents are driven maintaining a rigid formation towards the target, and multi-source scenario, in which each agent searches for the targets independently from the others.
\end{abstract}


\end{frontmatter}

\section{Introduction}
\subsection{Related work}
The target localization task has been widely studied in robotics. Gradient-based strategies, such as extremum seeking control, have been studied in~\cite{ariyur2003real} and applied in both single-agent scenarios~\cite{ZhangAGSK06} and multi-agent scenarios~\cite{switching-mas-grad}, \cite{mas-source-seeking}. In~\cite{schenato}, it is shown how a multi-agent formation can approximate the gradient of a scalar vector field, and a collaborative control law for source seeking and tracking is developed. However, gradient-based algorithms may remain stuck in local optima without finding the global solution. To overcome this issue, stochastic optimization algorithms have been developed, such as particle swarm optimization (PSO)~\cite{6878112}, ant colony optimization (ACO)~\cite{colorni1991distributed}, grey wolf optimization (GWO)~\cite{MIRJALILI201446} and gravity search optimization (GSA)~\cite{RASHEDI20092232}. These approaches mimic natural phenomena and implement stochasticity in order to avoid getting stuck in local, suboptimal solutions.

Acoustic source localization is a trending topic in robotics, specifically in the context of search and rescue applications. Exploiting sound information allows a robotic agent to locate a target even in challenging environmental conditions. Several algorithms have been developed to solve the problem of localizing sound sources. Multiple signal classification (MUSIC) exploits the orthogonality between the acoustic signal and noise subspaces to estimate the direction of arrival (DoA) of multiple sound sources~\cite{1143830}. However, the accuracy of the MUSIC algorithm degrades when the noise power is higher than the target. Many variants have been proposed to improve robustness against noise. In~\cite{6696920}, the generalized eigenvalue decomposition MUSIC (GEVD-MUSIC) introduces the noise correlation matrix for whitening the noise subspace. An enhanced version, called iterative GEVD-MUSIC (iGEVD-MUSIC) has been proposed in~\cite{6385994}. To reduce the algorithm's computational cost, generalized singular value decomposition MUSIC (GSVD-MUSIC) and iterative GSVD-MUSIC (iGSVD-MUSIC) have been presented in~\cite{8206494}. In~\cite{8206494}, a UAV embedded with a microphone array and guided by a human operator uses iGSVD-MUSIC and online robust principal component analysis (ORPCA) to detect a single frequency sound (whistle) in an outdoor environment. Other approaches to deal with robot noise (in particular drone's self-noise) have been studied. In~\cite{denoise-autoencoder}, a denoising autoencoder based on fully convolutional neural networks has been discussed. A general review on noise reduction for both ground and aerial platforms is reported in~\cite{noise-review}. One of the shortcomings of the MUSIC algorithm is that it provides only the direction of arrival of an acoustic signal, but no information regarding the distance from the source. In~\cite{article}, GEVD-MUSIC and iGEVD-MUSIC with distance estimation have been implemented on a UAV for search and rescue missions. Another strategy for sound source localization consists of estimating the time difference of arrival (TDoA) between different microphone pairs. The most popular TDoA-based method is the generalized cross-correlation with phase transformation (GCC-PHAT)~\cite{1162830}. A cross-correlation TDoA-based method with angular spectrum subtraction is investigated in~\cite{9670693} for a low signal-to-noise ratio environment. There also exist beamforming-based techniques for DoA estimation. However, these algorithms assume that the sound sources are located in the far field with respect to the microphones, and their performance degrades closer to the target~\cite{ARGENTIERI201587}.

\subsection{Problem Description}
Differently from the reviewed literature, in this work we assume that the agents can sense the environment only while they are not moving. We therefore design a switching mode controller that alternates between a listening phase, in which the agents estimate the DoA of the sound source exploiting recursive Bayesian estimation (RBE), and a movement phase, in which the agents drive towards the target. 
Indeed, this strategy reduces robots' self noise since there is no actuation during the listening and estimation phase, favoring the localization procedure. 

We adopt this scheme in two different scenarios: in the first case, a single target needs to be detected, and the agents drive the centroid of a rigid formation towards the source. In the second case, we assume that multiple incoherent sound sources need to be detected, and the agents search independently from each other, only communicating the localized targets positions. Throughout this work, it is assumed that the sources to be detected are static and undistinguishable from each other.

\section{Switching Mode Controller}
\subsection{Modeling}

Each agent's dynamics is described by double integrators equations:
\begin{equation}
    \ddot{p}_i = u_i \label{double_int}
\end{equation}
where $p_i = p(v_i) = (x_i, y_i)^T \in \mathbb{R}^2$ is the position of agent $v_i$ and $u_i$ is its control input, with $i = 1,...,N$, and $N$ is the total number of agents. We assume that each agent is equipped with a circular array of six omnidirectional microphones. Given the radius $r$ of the array, the position of the $h$-th microphone is given by
\begin{equation}
    p_h = p_i + r (\cos{(\frac{\pi}{3}(h-1) )}, \sin{( \frac{\pi}{3}(h-1) )})^T
    \label{mic_array}
\end{equation}

The acoustic intensity $I$ measured by a microphone is assumed to be constant if the distance from the microphone to the source is smaller or equal to $1m$, otherwise it is decreasing quadratically with the distance from the source:
\begin{equation}
    \left\{
        \begin{aligned}
            I &= \frac{W_0}{4\pi} &\quad& d \leq 1 \\
            I &= \frac{W_0}{4 \pi d^2} &\quad& d > 1
        \end{aligned}
    \right.
\end{equation}
where $W_0$ is the emission power of the source.

\subsection{Recursive Bayesian Estimation}
The core assumption of this work is that each agent can listen to the environment only while it is not moving. Therefore, we model our system as a stochastic hybrid system switching between a \textit{listening mode}, where the agents acquire measurements and estimate the sound DoA and the length to be covered in order to get closer to the source, and \textit{movement mode}. The procedure to estimate the step length and the sound DoA will be explained in the next section. To include the effect of background noise in our model, we assume that the step distance $s$ and DoA $\theta$ computed by the agents are sampled according to
\begin{subequations}
    \begin{align}
        \Tilde{s} \sim \mathcal{N}(s,\sigma_d^2) \\
        \Tilde{\theta} \sim vonMises(\theta, k_{\theta})
    \end{align}
\end{subequations}
The von Mises probability density function is defined as
\begin{equation}
    p(\theta) = \frac{1}{2 \pi I_0(k_{\theta})} exp(k_{\theta}(\cos{\theta-\mu}))
\end{equation}
where $I_0(k_{\theta})$ is the modified Bessel function of the first kind and order zero, $\mu$ is the expected value, and the parameter $k_{\theta}$ can be assimilated to the reciprocal of the variance of a normal distribution, $k_{\theta} \sim \sigma_d^{-2}$. 
Since while acquiring data the agents are still, the sequence of measurements can be modeled as a Markovian process, therefore, it is possible to obtain an estimate of the step length $\mu_s$ and its variance $P$ via RBE. Exploiting the fact that the product of two Gaussian distributions is proportional to a Gaussian distribution, the update equations between $k$ and $k+1$ for the step distance variance and mean can be written as
\begin{subequations}
    \begin{align}
        P_{k+1} = \frac{\sigma_d^2}{\sigma_d^2 + P_{k}} P_{k} \label{step_var}\\
        \mu_{s,k+1} = \frac{\Tilde{s}_{k+1} P_{k} +  \mu_{s,k}} {\sigma_d^2 + P_{k}} \label{step_mean}
    \end{align}
\end{subequations}
A similar reasoning holds for the recursive update equations of the DoA variance and mean value, considering that the convolution of two von Mises distribution is well approximated by a von Mises distribution~(\cite{vonMises}):
\begin{subequations}
\begin{align}
   K_{k+1} = & \bigl( (K_{k}\cos{\mu_{\theta,k}} + k_{\theta} \cos{\Tilde{\theta}_{k}})^2 + \\
             & (K_{k}\sin{\mu_{\theta,k}} + k_{\theta} \sin{\Tilde{\theta}_{k}})^2 \bigr)^{\frac{1}{2}}\nonumber \label{vonmises_var}
\end{align}
\begin{align}
   \mu_{\theta,k+1} = & ATAN2(K_{k} \sin{\mu_{\theta,k}} + k_{\theta} \sin{\Tilde{\theta}_{k}}, \\
             & K_{k} \cos{\mu_{\theta,k}} + k_{\theta} \cos{\Tilde{\theta}_{k}}) \nonumber \label{vonmises_mean}
\end{align}
\end{subequations}

\subsection{Switching Conditions and Reset} \label{conditions}
$P$ and $K^{-1}$ represent the uncertainty of the estimates of the step length and the  DoA, respectively. Considering \eqref{step_var}, it is easy to notice that the variance is monotonically decreasing with the number of iterations. 
The DoA variance update equation can be rewritten as
\begin{equation}
    K_{k+1} = \sqrt{K_k^2 + k_{\theta}^2 + 2 K_k k_{\theta} \cos{(\mu_{\theta,k}-\Tilde{\theta}_k)}} \label{vonmises_var2}
\end{equation}
and, assuming $-{\frac{\pi}{2}} \leq \mu_{\theta,k}-\Tilde{\theta}_k \leq {\frac{\pi}{2}}$, this quantity is also decreasing with the number of iterations. 
Therefore, under this assumption, an intuitive choice for the \textit{measurement-to-movement} switching condition is to wait until the estimation is precise enough, namely
\begin{subequations}
    \begin{align}
        P_k \leq P_{thresh} \\
        K_k^{-1} \leq K_{thresh}
    \end{align}
    \label{meas2move}
\end{subequations}
Equation \ref{step_var} shows that the speed of convergence for the step distance estimator depends on the parameter $\sigma_d^2$. In fact, $\lim_{\sigma_d^2 \to 0} P_{k+1} = 0$ and the variance decreases to $0$ almost instantly. On the other hand, $\lim_{\sigma_d^2 \to \infty} P_{k+1} = P_k$, therefore the variance is constant and the threshold may never be reached. 
A similar reasoning can be carried out for the DoA variance estimator, considering the expression in \eqref{vonmises_var2}. In particular, it holds $K_{k+1} \leq K_k + k_{\theta}$. Then, considering $\lim_{k_{\theta} \to 0} K_{k+1} = K_k$ and convergence may never be reached. In general, higher variances negatively affect the speed of convergence of the RBE estimators. 

Considering the \textit{movement-to-measurement} switching condition, the switch happens when the system has traveled a distance longer than or equal to the step distance in \eqref{step_mean}:
\begin{equation}
    \Vert p - p_{meas} \Vert \geq \mu_{s} \label{mov2meas}
\end{equation}
where $p_{meas}$ is the position during the measurement mode (constant), and $p$ is the current position while in movement mode. How the step $\mu_s$ is computed will be explained in the next section, however, in order to reach the target, it is sufficient that the step decreases while the distance from the source is decreasing. Moreover, whenever the system switches from movement mode to measurement mode, a reset of the estimators is performed, by initializing the variances as $P_0 \rightarrow \infty$, $K_0 = 0$. This allows the system to recover from bad estimates or from initial states that are not meaningful with respect to the real ones. Indeed, by performing this initialization, it holds
\begin{subequations}
    \begin{align}
        \mu_{s,1} \approx \Tilde{s}_1 \\
        \mu_{\theta,1} \approx \Tilde{\theta}_0
    \end{align}
\end{subequations}

where $\Tilde{s}_1$ and $\Tilde{\theta}_0$ are respectively the noisy step distance and DoA computed at the first estimation iteration while in measurement mode.

\section{Single Source Scenario}
In the first scenario, we assume that a single static source needs to be detected. The goal is to drive the centroid of a multi-agent formation as close as possible to the target.
Without loss of generality, and to simplify the discussion, hereafter the scenario is considered as a 2D environment.
\subsection{Bearing Rigid Formation}
 We model the formation of agents as a bearing rigid framework $(\mathcal{G},p)$ (\cite{michieletto2021unified}), where $\mathcal{G} = (\mathcal{V}, \mathcal{\xi})$ is a graph with a set of nodes $\mathcal{V}$ of cardinality $\vert \mathcal{V} \vert = N$ and a set of undirected edges $\mathcal{\xi}$ of cardinality $\vert \mathcal{\xi} \vert = E$, while $p: \mathcal{V} \rightarrow \mathbb{R}^2$ maps each node of the graph to a point in the 2D plane. The inter-agent bearing vector is defined as 
\begin{equation}
    b_{i,j} = \frac{p_j - p_i}{\Vert p_j - p_i \Vert} \in \mathbb{S}^1
\end{equation}
where $v_i$, $v_j \in \mathcal{V}$. Moreover, it is assumed that each agent in the formation is equipped with an omnidirectional microphone. This can be achieved with the microphones described in \eqref{mic_array} by taking the mean of the measurements of each channel in the array.

\subsection{DoA and Step Distance Estimation}
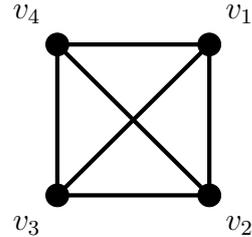
\begin{figure}[hb]
    \centering
    \begin{tikzpicture}
        \draw[ultra thick] (-1,0) -- (1,0);
        \draw[ultra thick] (1,0) -- (1,-2);
        \draw[ultra thick] (-1,-2) -- (1,-2);
        \draw[ultra thick] (-1,0) -- (-1,-2);
        \draw[ultra thick] (-1,-2) -- (1,0);
        \draw[ultra thick] (-1,0) -- (1,-2);
        \filldraw[black] (-1,0) circle (0.15);
        \node at (1.4, 0.4) {\large $v_1$};
        \filldraw[black] (1,0) circle (0.15);
        \node at (1.4, -2.4) {\large $v_2$};
        \filldraw[black] (1,-2) circle (0.15);
        \node at (-1.4, -2.4) {\large $v_3$};
        \filldraw[black] (-1,-2) circle (0.15);
        \node at (-1.4, 0.4) {\large $v_4$};
    \end{tikzpicture}
    \caption{Four agents bearing rigid formation}
    \label{formation_fig}
\end{figure}

In order to detect a sound-emitting source in $\mathbb{R}^2$, at least three non-collinear microphones are required. This is easily achieved with a bearing rigid formation. In our work, we consider a bearing rigid square formation of four agents ($v_1$, $v_2$, $v_3$, $v_4$) as in Figure~\ref{formation_fig}.

It is possible to obtain a first-order approximation of the intensity directional derivative along the direction defined by the inter-agent bearing as
\begin{equation}
    \nabla_{b_{i,j}} I = \Delta I_{i,j} b_{i,j}
\end{equation}
where $\Delta I_{i,j} = I_j - I_i$, and $I_i$, $I_j$ represent the acoustic intensity perceived by agents $i$ and $j$, respectively. Differently from \cite{schenato}, we approximate the maximum ascent direction by considering a linear combination of such directional derivatives defined with respect to a subset $\xi^{\star} \subseteq \xi$:
\begin{equation}
    \nu_{\theta} = \sum_{i,j \in \xi^{\star}} \Delta I_{i,j} b_{i,j}
\end{equation}
and finally we obtain $\theta = ATAN2(\nu_{\theta,y},\nu_{\theta,x})$. This approximation is independent from the formation geometry, the only requirement is that vectors in $\xi^{\star}$ span the space $\mathbb{R}^2$. Concerning the step estimation, as highlighted in the previous section, this quantity should be inversely proportional to the real distance from the source, in order to avoid overshooting the target. We choose to compute this quantity as
\begin{equation}
    s =  max \left\{ \alpha \left| \frac{1}{I_1} - \frac{1}{I_3} \right|, \alpha \left| \frac{1}{I_2} - \frac{1}{I_4} \right| \right\}
\end{equation}
with $\alpha \in \mathbb{R}$, $ \alpha > 0$, since the intensity increases as the distance to the source decreases.

\subsection{Bearing Maneuvering Control} 
In order to lead the formation to the target while maintaining the desired bearing rigid formation, we implement the bearing maneuver control in~\cite{7307285}, where at least two agents are elected leaders and follow a trajectory $p_{i,d}$ designed to reduce the centroid-target distance, while the other agents are required to follow the leaders while maintaining the desired bearing formation. The trajectory for the leaders is defined in terms of velocity as
\begin{equation}
    \Dot{p}_{i,d} = c (\cos{\mu_{\theta}}, \sin{\mu_{\theta}})^T
\end{equation}
where $\mu_{\theta}$ is the estimated DoA and $c$ represents a constant body frame velocity. For the leaders, a simple PD controller is adopted:
\begin{equation}
    u_i = K_d \Dot{e}_i + K_p e_i \label{leaders_control}
\end{equation}
where $e_i = p_{i,d} - p_i$, while for the followers the control law is
\begin{equation}
    u_i = -\sum_{j \in \mathcal{N}(i)} P_{b_{i,j}^{d}} (k_d (\Dot{p}_i - \Dot{p}_j)) + k_p (p_i - p_j))
    \label{followers_control}
\end{equation}
where $\mathcal{N}(i)$ represents the set of neighbors of node $i$, $b_{i,j}^d$ is the desired bearing between agents $i$ and $j$ and $P_{b_{i,j}^{d}} = I(2) - b_{i,j}^{d} (b_{i,j}^{d})^T$ is a projector onto the space orthogonal to $b_{i,j}^{d}$.

\section{Multiple Sources Scenario}
\subsection{DoA and Step Distance Estimation}
We now consider an environment with multiple, undistinguishable sources to be detected. 
In order to ensure better coverage of the search area, each agent searches independently from the others, exploiting the information from the six channels of the microphone array. The DoA is computed as $\theta = ATAN2(\nu_{\theta,y}, \nu_{\theta,x})$, where
\begin{equation}
    \nu_{\theta} = \frac{p(I_{MAX}) - p(I_{min})}{\Vert p(I_{MAX}) - p(I_{min}) \Vert}
\end{equation}
where $p(I_{MAX})$ and $p(I_{min})$ are the positions of the microphones reading the maximum and minimum intensity, respectively. The step distance is instead computed according to
\begin{equation}
    s = \beta \left( \frac{1}{I_{min}} - \frac{1}{I_{MAX}} \right)
\end{equation}
where $\beta \in \mathbb{R}$, $\beta > 0$ is a scaling parameter. 
Differently from the single source scenario, in this case, the agents are working independently; therefore, the switching conditions in \eqref{meas2move} and \eqref{new_mov2meas} are applied individually to each robot, allowing agents to operate in different modes independently rather than being constrained to the same operating mode.

\subsection{Exploration Algorithm and Control}
Each source represents a basin of attraction for the agents since they follow a gradient-based direction. In order to allow the agents to detect different sources, they must be able to explore and escape from such basins of attraction. We therefore define, for each $i$-th agent, a virtual velocity $\Bar{v}_i$, and we modify the \textit{movement-to-measurement} switching condition in \eqref{mov2meas} to become
\begin{equation}
    \Vert p_i - p_{meas,i} \Vert \geq \gamma \Vert \Bar{v}_i \Vert \label{new_mov2meas}
\end{equation}
where $p_{meas,i}$ is the position of agent $i$ in measurement mode (constant), $p_i$ is the position of agent $i$ while in movement mode and $\gamma = 1s$. The virtual velocity is initialized as $\Bar{v}_i(0) = \mu_{s,i} (\cos{\mu_{\theta,i}}, \sin{\mu_{\theta,i}})^T$ and it is updated each time the agent switches from measurement mode to movement mode. 

To guarantee exploration, each robot needs to define an explored area that contains a detected target and that needs to be avoided by itself and the other agents. We assume that a target is detected by agent $i$ if the step estimated by that agent is smaller than a given threshold, $\mu_{s,i} \leq \mu_{s,thresh}$. We approximate the detected target position $p_t$ as the position of the detecting agent $p_i$ and we define the explored area as a circular area of center $p_t = p_i$ and radius $r_t$. Whenever an agent meets the detection condition, if its position lies inside an already existing explored area, the radius of such area is increased according to $r_{t,k+1} = k_r r_{t,k}$, with $k_r>1$. Otherwise, a new area of radius $r_{t+1,0} = r_0$ is defined. We assume that the union of all explored areas is shared between the agents. Then, the virtual velocity for each agent can be updated as
\begin{equation}
 \Bar{v}_{i,k+1} =
    \begin{cases}
        k_r^{\prime} r_{t,k}\begin{bmatrix}\cos{\theta_{rd}}\\ \sin{\theta_{rd}}\end{bmatrix},  \exists p_t, r_t\;\text{s.t.}\; \Vert p_i - p_t \Vert \leq r_t \\
      k_v \Bar{v}_{i,k} + \mu_{s,i} \begin{bmatrix}\cos{\mu_{\theta,i}} \\ \sin{\mu_{\theta_i}}\end{bmatrix},\: \text{otherwise}
    \end{cases}
    \label{virtual_v}
\end{equation}
where $k_v$, $k_r^{\prime}$ $\in \mathbb{R}$, $ 0 < k_v < 1$, $k_r^{\prime} \geq 1$, the vector $(\cos{\theta_{rd}}, \sin{\theta_{rd}})^T$ represents a random direction spanning the circle at intervals of $\frac{\pi}{4}$. The trajectory for agent $i$ is then defined as 
\begin{equation}
    \Dot{p}_{i,d} = c (\cos{\theta_d}, \sin{\theta_d})^T
\end{equation}
where $\theta_d = ATAN2(\Bar{v}_{i,y}, \Bar{v}_{i,x})$. The control law for each agent is a PD control as in \eqref{leaders_control}. It can be noticed that, in \eqref{virtual_v} for the case where the agent is inside an explored area, the virtual velocity norm is always greater than the current explored area radius. This fact, according to the switching condition in \eqref{new_mov2meas}, ensures that the agent exits the already visited area. Moreover, the closer the decay factor $k_v$ is to $1$, the longer the random directions taken when inside an explored area affects the agent's trajectory, therefore favoring exploration.
\begin{figure*}[t]
    \centering
    \begin{minipage}{.32\textwidth}
        \includegraphics[scale = .46]{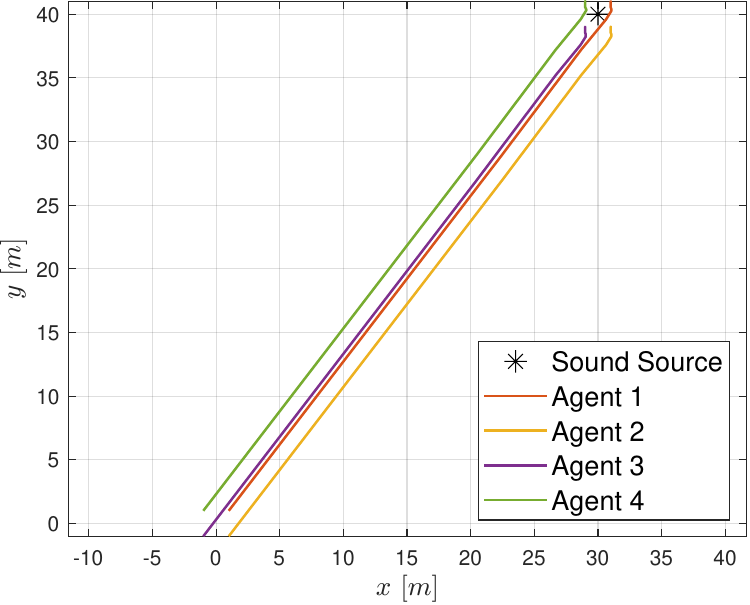}
        \caption{Formation behaviour ($\sigma_d^2 = 0.1$, $k_{\theta} = 1$)}
        \label{formation}
    \end{minipage}
     \begin{minipage}{.32\textwidth}
        \includegraphics[scale = .46]{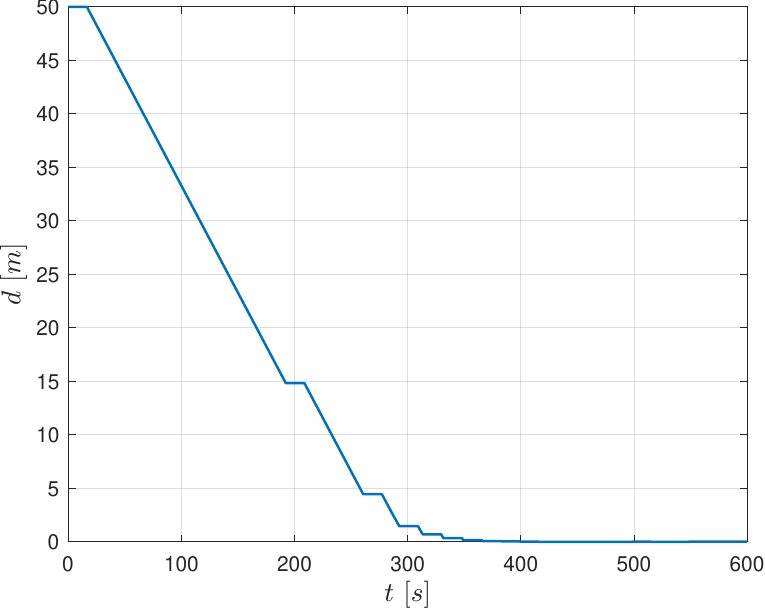}
        \caption{Formation's centroid distance from the target ($\sigma_d^2 = 0.1$, $k_{\theta} = 1$)}
        \label{centroid_dist}
    \end{minipage}
     \begin{minipage}{.32\textwidth}
        \includegraphics[scale = .46]{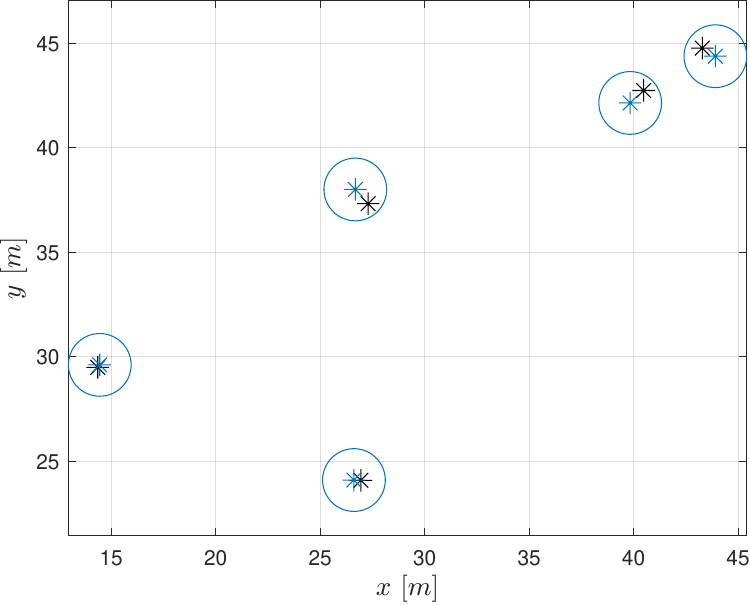}
        \caption{Real targets (in black) and detected targets (in blue)}
        \label{detections5}
    \end{minipage}
\end{figure*}

\section{Simulations and Results}
To validate the algorithms presented in this work, we perform some tests in a simulated environment for both the single source scenario and the multiple source scenario. For these simulations, the hybrid dynamics is approximated by considering the discretization of the agents models in \eqref{double_int} and of the control laws \eqref{leaders_control}-\eqref{followers_control}, with a sample time $T_s = 1ms$. For all the simulations, we assume that each source's power is $W_0 = 10^8 W$.

\subsection{Single Target Simulations}
We test the RBE estimator's robustness against different level of noise. We consider a single target with its position fixed at $p_T = (30m,40m)^T$. The bearing rigid formation is the one reported in Figure~\ref{formation_fig}, with the agent's starting positions at $p_1(0) = (1m,1m)^T$, $p_2(0) = (1m,-1m)^T$, $p_3(0) = (-1m,-1m)^T$ and $p_4(0) = (-1m,1m)^T$, so that the centroid of the formation is located at $p_c(0) = (0m,0m)^T$. 
The set of inter-agent bearings considered for the DoA approximation is $\xi^{\star} = \{ b_{21}, b_{41} \}$, and we consider agents $1$ and $3$ as the leaders. The control parameters are set as $K_p = K_d = k_p = k_d = 10$, and the reference velocity for the agents is $c = 0.2 m/s$. The thresholds for the \textit{measurement-to-movement} switching conditions are set as $P_{thresh} = K_{thresh} = 1 \times 10^{-4}$, while the scaling factor $\alpha$ is set to $10^6$. 

The goal of these tests is to compare the time required for the formation to converge to the target with respect to different noise variances. The time of convergence $t_s$ is defined as the minimum time for which the system is in measurement mode, and the distance $d$ of the centroid from the target is $d \leq 0.05m$ $\forall t \geq t_s$. 
\renewcommand{\arraystretch}{1.1}
\begin{table}[hb]
    \begin{center}
    \caption{Time of convergence with respect to different levels of variances}
    \begin{tabular}{|c|c|c|c|}
         \hline
         \diagbox{$\sigma_d^2$}{$k_{\theta}$} & $100$ & $10$ & $1$ \\
         \hline
         $0.01$ & $251.662$ & $259.034$ & $382.876$ \\
         
          $0.1$ & $258.768$ & $259.077$ & $382.870$ \\
          
          $1$ & $340.709$ & $340.724$ & $382.850$ \\
          
          $10$ & $1050.764$ & $1050.770$ & $1050.794$ \\
          
          $100$ & $9250.778$ & $9250.773$ & $9250.756$ \\
          \hline
    \end{tabular}
    \label{tab_convergence}
    \end{center}
\end{table} \\
\renewcommand{\arraystretch}{1}
Table \ref{tab_convergence} reports the time of convergence $t_s$ (in seconds) against different values for the DoA and step distance variances. It can be noticed that, for increasing values of the variances, the formation takes longer to reach the target. Since the estimators are independent from each other and the switching condition must be met by both the variances in order to cause the system to switch its operating mode, the convergence time is dictated by the slower estimator. Moreover, we have observed that for $\sigma_d^2 = 0.01$, $k_{\theta} = 0.1$, the centroid of the formation does not move for $1.5 \times 10^{4}s$, meaning that for this time the DoA estimator did not reach the desired variance. Those results are coherent with the reasoning in section \ref{conditions}. 

Figure \ref{formation} shows the agents moving towards the target maintaining the desired formation, while in Figure \ref{centroid_dist}, the centroid distance from the target can be observed. The intervals for which the distance is constant correspond to the time in which the formation is in listening mode. It can also be noticed that the distance traveled by the centroid decreases as the formation gets closer to the source.

\subsection{Multiple Targets Simulations}
We now show some results concerning the multiple sources scenario, by testing the detection ability of the algorithm. 
The targets are randomly spawned within a square area of $50m^2$, while the agents start at the corners of a square of length $60m$, therefore they are located outside the search area. This is done to avoid lucky initialization scenarios. The sound sources are assumed to be incoherent, therefore the overall intensity is given by the sum of the intensities of each source. The estimators thresholds are defined as in the single target scenario, as well as the robot velocity $v_c$. The scaling factor $\beta$ is chosen as $\beta = 4 \times \sqrt{10^{13}}$. When a new target is detected, the initial radius for the explored area is set at $r_{t,0} = 3.1m$, and it is increased by $20\%$ whenever the same target is detected. The decay factor for the robot's virtual velocity is chosen as $k_v = 0.9$, while the coefficient $k_r^{\prime}$ is set to $1.1$. 

We consider variances $\sigma_d^2 = 0.01$ and $k_{\theta} = 100$. We consider cases where the number of targets varies between three and eight, while keeping four seeker agents. For each case, the average number of detected sources over ten simulations of $1000s$ is computed. In this work, we assume that, knowing the location of a target, the rescue team searches in a circular area of radius $r_{tt} = 1.5m$ around the detected target. Therefore, we consider a source to be located if it lies inside a circle of radius $r_{tt}$ around the point detected by an agent.
\renewcommand{\arraystretch}{1.1}
\begin{table}[h]
    \begin{center} 
    \caption{Average number of detected sources}
    \begin{tabular}{|c|c|}
       \hline
       \textbf{Number of targets}  & \textbf{Average number of detections} \\
       \hline
        $3$ & $2.7$ \\
        $4$ & $3.6$ \\
        $5$ & $4.4$ \\
        $6$ & $4.5$ \\
        $7$ & $5.4$ \\
        $8$ & $6.4$ \\
        \hline
    \end{tabular}
    \label{avg_detections}
    \end{center}
\end{table} \\
\renewcommand{\arraystretch}{1}
Table~\ref{avg_detections} shows the average number of detected sources with respect to the number of targets. It can be noticed that the algorithm's performance decreases with the number of targets. Several reasons may be the cause of targets being missed by the robots: first, each simulation has a fixed length of $1000s$, however it is possible that, with more time, the agents are able to detect more sources. Moreover, there is a random element in the generation of the trajectory for each agent, when they are inside an already explored area, therefore it cannot be guaranteed that an agent drives towards an unknown target. Figure \ref{detections5} shows an example of the detections performed by the agents.

\section{Conclusion}
A novel multi agent stochastic hybrid system for sound based target localization has been presented, and proofs of concept to demonstrate the effectiveness of the proposed algorithms have been developed considering a simplified environment.

Concerning the single target case, we showed how the proposed switching mode control can precisely locate a source even in the presence of high noise variance. However, according to the reasoning in section \ref{conditions}, for extremely high noise variances the estimators may fail to converge. Considering the multi-target case, stochasticity has been introduced to embed exploration capabilities in the agents, allowing a single robot to detect many sources. Although stochasticity and increasing the explored area radius is helpful for exploration, it may increase the time needed to detect all the targets in the environment, or, in some cases, prevent the agents from finding a particular source. 

Studying the effectiveness of this strategy in more realistic, complicated, and possibly dynamic environment, as well as considering data fusion approaches combining data from different sensors, such as cameras and thermal sensor, will be the goal of future developments.

\bibliography{ref}             
                                                   






\end{document}